\documentclass[runningheads]{llncs}

\usepackage{iciap}
\usepackage{iciapabbrv}

\usepackage[T1]{fontenc}
\usepackage{tikz}
\usepackage{xspace}
\usepackage{xcolor}
\usepackage{graphicx}
\usepackage{booktabs}
\usepackage{makecell}
\usepackage{colortbl}
\usepackage{amsmath,amssymb,amsfonts}
\usepackage[accsupp]{axessibility}
\usepackage{hyperref}
\usepackage{orcidlink}

\definecolor{blue}        {HTML}{4169E3}
\definecolor{gold}        {HTML}{FBF2D2}
\definecolor{green}       {HTML}{147546}
\definecolor{silver}      {HTML}{DDDDDD}
\definecolor{bronze}      {HTML}{EED2B8}
\definecolor{gold_dark}   {HTML}{D9AE13}
\definecolor{silver_dark} {HTML}{909090}
\definecolor{bronze_dark} {HTML}{9A5F26}

\def\ours{\textbf{KairosAD}\xspace}
\newcommand{\medal}[3]{%
    \tikz[baseline=(char.base)]{%
        \node[rounded corners=2pt,fill=#1,draw=#2,inner sep=1.5pt] (char) {#3};%
    }%
}
\newcommand{\bm}[2]{%
    \ifnum#1=1
        \medal{gold}{gold_dark}{\textbf{#2}}%
    \else\ifnum#1=2
        \medal{silver}{silver_dark}{#2}%
    \else\ifnum#1=3
        \medal{bronze}{bronze_dark}{#2}%
    \else
        #2%
    \fi\fi\fi
}

\begin{document}

\title{KairosAD: A SAM-Based Model for Industrial Anomaly Detection on Embedded Devices} 

\titlerunning{KairosAD: Industrial Anomaly Detection on Embedded Devices}

\author{
Uzair Khan\orcidlink{0000-0003-4107-4359} \and
Franco Fummi\orcidlink{0000-0002-4404-5791} \and
Luigi Capogrosso\orcidlink{0000-0002-4941-2255}
}

\authorrunning{U.~Khan et al.}

\institute{Department of Engineering for Innovation Medicine, University of Verona, Italy\\
\email{name.surname@univr.it}}

\maketitle

\begin{abstract}
In the era of intelligent manufacturing, anomaly detection has become essential for maintaining quality control on modern production lines.
However, while many existing models show promising performance, they are often too large, computationally demanding, and impractical to deploy on resource-constrained embedded devices that can be easily installed on the production lines of Small and Medium Enterprises (SMEs).
To bridge this gap, we present \ours{}, a novel supervised approach that uses the power of the Mobile Segment Anything Model (MobileSAM) for image-based anomaly detection.
\ours{} has been evaluated on the two well-known industrial anomaly detection datasets, \emph{i.e.}, MVTec-AD and ViSA.
The results show that \ours{} requires 78\% fewer parameters and boasts a 4$\times{}$ faster inference time compared to the leading state-of-the-art model, while maintaining comparable AUROC performance.
We deployed \ours{} on two embedded devices, the NVIDIA Jetson NX, and the NVIDIA Jetson AGX.
Finally, \ours{} was successfully installed and tested on the real production line of the Industrial Computer Engineering Laboratory (ICE Lab) at the University of Verona.
The code is available at \url{https://github.com/intelligolabs/KairosAD}.

\end{abstract}

\section{Introduction} \label{cha:intro}

Image-based anomaly detection of industrial components plays a crucial role in enabling fully automated production and ensuring optimal quality control in manufacturing.
In particular, image anomaly detection can be broadly classified into \emph{semantic anomaly detection} and sensory anomaly detection~\cite{yang2024generalized}.
Semantic anomaly detection emphasizes the overall semantics of the image, while sensory anomaly detection focuses on fine-grained details, such as minor scratches on objects.
This article concentrates on semantic anomaly detection, a task commonly performed in Small and Medium Enterprises (SMEs), where it plays a crucial role in maintaining production efficiency and ensuring product quality~\cite{capogrosso2024diffusion,girella2024leveraging}.

Recently, several approaches, based on teacher-student networks~\cite{deng2022anomaly}, reconstructions~\cite{liang2023omni}, or memory banks~\cite{roth2022towards}, have been developed.
Most of these anomaly detection methods are highly effective in classifying data into normal (fault-free) or anomalous categories.
However, many of these architectures are large and computationally intensive.
They demand significant processing power and memory, making them unsuitable for deployment on embedded devices~\cite{li2024sam}.

However, SMEs, which often have limited infrastructure and budgets, often rely on low-cost edge computing devices for real-time quality control on their production lines~\cite{khan2025comprehensive}.
This reliance on less powerful hardware creates a gap between the high performance of advanced anomaly detection models and their practical deployability in real-world manufacturing settings.
As a result, many SMEs struggle to adopt state-of-the-art techniques due to constraints in computational resources and integration complexity.

\begin{figure*}[t!]
    \centering
    \includegraphics[width=\linewidth]{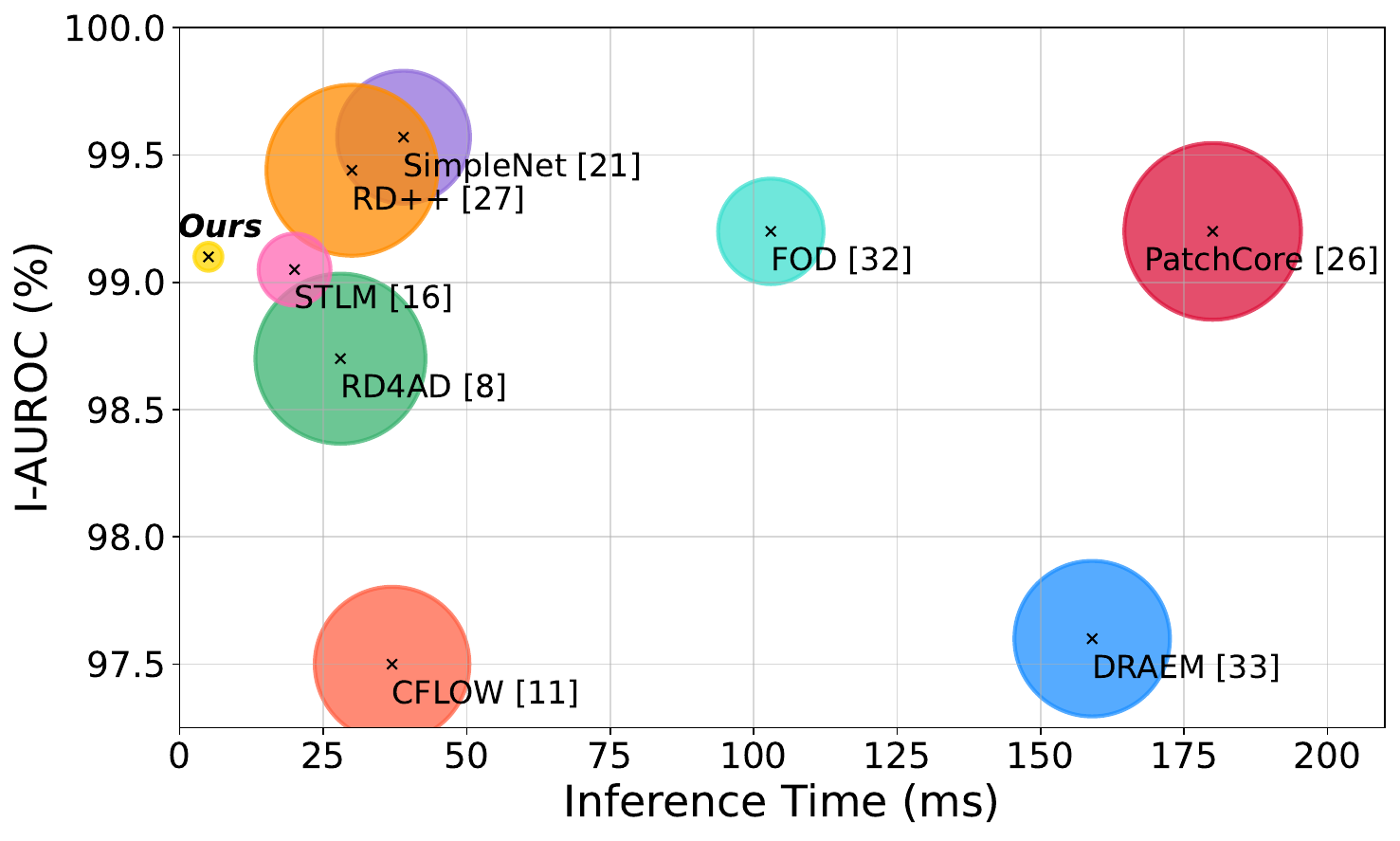}
    \caption{Comparisons of different anomaly detection methods in terms of I-AUROC (vertical axis), inference time (horizontal axis), and the ratios of parameter numbers (circle radius), on the MVTec-AD dataset~\cite{bergmann2019mvtec}.
    Our \ours{} achieves competitive I-AUROC results, while having 78\% fewer parameters with respect to SimpleNet~\cite{liu2023simplenet}, \emph{i.e.}, the model that achieves the highest AUROC.
    Instead, with respect to STLM~\cite{li2024sam}, which represents the state-of-the-art of efficient deep learning method for anomaly detection, \ours{} has 35\% of fewer parameters, achieves 4$\times{}$ speedup performance in terms of inference time, with an increase in I-AUROC performance as well.}
    \label{fig:fig_1}
    \vspace{-1em}
\end{figure*}

To address this challenge, we introduce \emph{\ours{}}, a novel supervised image-level anomaly detection model.
\ours{} leverages the power of \emph{Mobile Segment Anything Model (MobileSAM)}~\cite{zhang2023faster} to achieve accurate anomaly detection performance with a significantly reduced computational cost.
Unlike existing methods that often require complex architectures and large parameter counts, \ours{} is designed for simplicity and efficiency.
In fact, as we can see from Figure~\ref{fig:fig_1}, \ours{} effectively balances precision, speed, and number of parameters, making it ideal for real-time deployment on embedded devices.

We evaluated \ours{} on two widely used industrial anomaly detection datasets: \emph{MVTec-AD}~\cite{bergmann2019mvtec} and \emph{ViSA}~\cite{zou2022spot}.
Furthermore, we deployed \ours{} on several embedded devices, such as \emph{NVIDIA Jetson NX} and \emph{NVIDIA Jetson AGX}, and successfully installed and tested on the real production line of the Industrial Computer Engineering Laboratory (ICE Lab) at the University of Verona.

Finally, \ours{} demonstrates remarkable \emph{eco-friendliness}, consuming only 0.0465 kWh of energy and producing only 0.0154 kg of CO\textsubscript{2}eq emissions in both the training and testing phases\footnote{Results calculated by using: \url{https://mlco2.github.io/codecarbon/}.}.
Even its power consumption remains optimized, with an average GPU usage of approximately 270W during training, highlighting the model’s ability to achieve high performance while significantly reducing its environmental footprint compared to recent deep learning approaches.

\section{Related Work} \label{cha:related}

Over the years, anomaly detection in the industry domain has been extensively explored~\cite{liu2024deep} (Section~\ref{sec:industry_ad}).
Since \ours{} is based on MobileSAM and employs efficiency-driven methods to enable deployment on embedded systems, this section also delves into the literature on foundation models for vision (Section~\ref{sec:vision_foundation_models}) and efficient deep learning techniques (Section~\ref{sec:efficient_dl}).

\subsection{Industrial Anomaly Detection} \label{sec:industry_ad}
Although extensive research has been carried out on anomaly detection, industrial image data presents unique challenges~\cite{bergmann2019mvtec}.

Many industrial anomaly detection methods focus on image reconstruction and detect anomalies based on the reconstruction error~\cite{liu2024deep}.
Generative models such as autoencoders~\cite{bergmann2018improving}, variational autoencoders~\cite{liu2020towards}, and generative adversarial networks~\cite{akcay2019ganomaly} are the architectures most widely used to reconstruct normal images from anomalous ones.
Nonetheless, these methods face certain limitations, especially when reconstructing complex industrial textures and patterns.

Recent approaches leverage memory banks, where a core set of stored features from a pre-trained backbone is used to compute patch-level distances for anomaly detection~\cite{damm2024anomalydino}.
PatchCore~\cite{roth2022towards} is a foundational approach that extracts patch-level feature embeddings of normal images into a memory bank, detecting anomalous patches during inference through a patch query process.
SoftPatch~\cite{jiang2022softpatch}, based on PatchCore, introduced a patch-level filtering strategy in which patch features are filtered and weighted before being stored in the memory bank to reduce contamination from anomalous patches, thus enhancing model robustness.
However, most of these methods come at the cost of increased computational complexity and a large memory space, making them unsuitable for deployment on resource-constrained devices.

In terms of efficient industrial anomaly detection models, our main competitor is STLM~\cite{li2024sam}.
Unlike STLM, which employs a two-branch architecture, we adopt just a single-branch design, reducing complexity and improving inference speed.
Furthermore, while STLM focuses on both anomaly detection and localization, our method is specifically optimized for just the image-level task, prioritizing efficiency for real-time deployment.
From a methodological perspective, STLM is based on feature distillation and contrastive learning, while \ours{} uses MobileSAM for a lightweight yet effective feature extraction, resulting in a simpler and more computationally efficient model.

\subsection{Foundation Models for Vision} \label{sec:vision_foundation_models}
Multimodal foundation models have emerged as powerful tools across various tasks~\cite{li2024multimodal}.
For visual anomaly detection, particularly relevant are multimodal approaches based on CLIP~\cite{radford2021learning} and Segment Anything Model (SAM)~\cite{kirillov2023segment}, as well as vision-only models like DINO~\cite{caron2021emerging}. 

Specifically, CLIP learns visual concepts from natural language descriptions by training on image-text pairs using a contrastive learning objective that aligns embeddings from both modalities.
This shared feature space enables downstream applications, such as zero-shot image classification, by comparing image embeddings to class-specific textual prompts.

Instead, DINO adopts a self-supervised student-teacher framework based on Vision Transformers (ViT), leveraging a multiview strategy to predict softened teacher output, resulting in robust and high-quality feature representations.
DINOv2~\cite{oquab2023dinov2} extends these ideas by incorporating patch-level reconstruction techniques and scaling to larger architectures and datasets.

Beyond these, SAM and its efficient variant, such as MobileSAM~\cite{zhang2023faster}, have introduced a new paradigm in vision models, particularly for segmentation tasks.
SAM is designed as a powerful promptable segmentation model that generalizes well across diverse images, enabling accurate object delineation with minimal supervision.
MobileSAM builds upon SAM, optimizing it for edge devices by reducing computational demands while maintaining competitive performance.
These models, though primarily designed for segmentation, have shown potential for various downstream tasks, including anomaly detection.
As a result, we take advantage of their efficiency and adaptability to develop \ours{}, which is based on the MobileSAM image encoder.

\subsection{Efficient Deep Learning} \label{sec:efficient_dl}
Over the past decades, a large amount of research has been invested in improving embedded technologies to enable real-time solutions for many complex applications.
However, deploying learning models on tiny devices is substantially difficult due to severe architectural, energetic, and latency constraints~\cite{capogrosso2024machine}.

Several techniques aim to reduce the size and computational cost of the model without sacrificing performance.
These include pruning~\cite{vadera2022methods}, quantization~\cite{gholami2022survey}, and knowledge distillation~\cite{gou2021knowledge}.
Furthermore, in order to provide lightweight models capable of delivering acceptable performances for their intended applications, specialized techniques have been proposed in model architecture exploration, model simplification, and architectural modifications, such as Neural Architecture Search (NAS)~\cite{ren2021comprehensive}, and the attention mechanism~\cite{brauwers2021general}.

Although commonly associated with notable contributions to machine translation tasks, the attention mechanism has been adapted for a wide range of applications, allowing the model to selectively look for different features, facilitating the extraction of relevant information from complex and high-dimensional data~\cite{capogrosso2024machine}.
One of the most significant advances that utilize this principle is the Transformer architecture~\cite{vaswani2017attention}.

Recent efforts to develop computationally efficient Transformers in vision have broadened their potential use in resource-constrained settings~\cite{khan2022transformers}.
MobileSAM represents one of the latest advances, and for this reason, we decided to use the MobileSAM image encoder in \ours{}.

\section{Methodology}  \label{cha:method}

\begin{figure*}[t!]
    \centering
    \includegraphics[width=\linewidth]{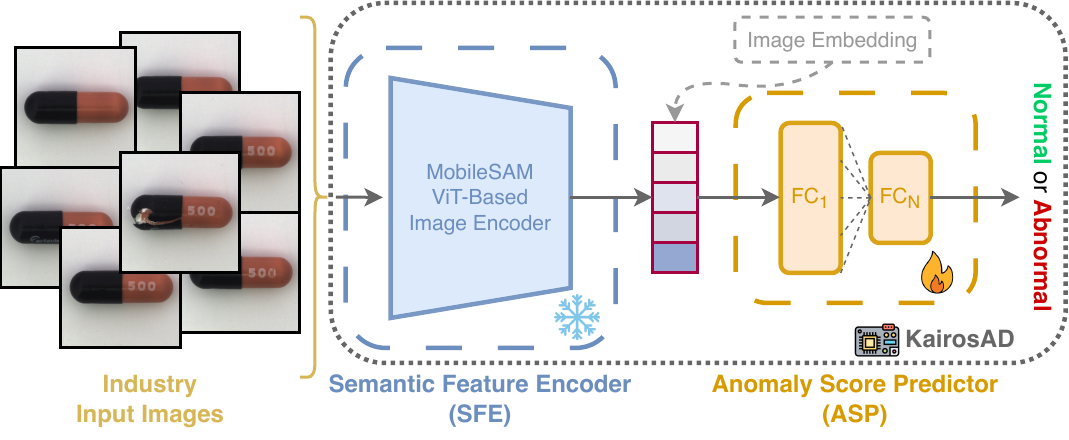}
    \caption{The \ours{} architetcure.
    Specifically, \ours{} is composed of a Semantic Feature Encoder (SFE) and an Anomaly Score Prediction (ASP).
    This simple but effective model allows us to perform efficient image-level anomaly detection on embedded devices.}
    \label{fig:fig_2}
\end{figure*}

In this section, we introduce \ours{}, a model designed for image-level industrial anomaly detection on embedded devices.
The overall architecture is shown in Figure~\ref{fig:fig_2}.

\subsection{The Semantic Feature Encoder (SFE) Sub-Network} \label{sec:sfe_subnetwork}
The SAM model is based on a Vision Transformer (ViT-H) with 632 million parameters to generate high-quality image embeddings.
However, this high computational cost makes the SAM encoder unsuitable for deployment on resource-constrained devices.

To address this limitation, MobileSAM~\cite{zhang2023faster} was introduced as a lightweight alternative, incorporating a more efficient image encoder with only 5.78 million parameters while maintaining comparable performance.
The transition from the high-parameter ViT-H encoder to the significantly more compact MobileSAM encoder is achieved through a well-structured knowledge distillation process.
In particular, instead of jointly optimizing both the encoder and the decoder, the authors separate the training of the image encoder from that of the mask decoder.
This approach allows the lightweight encoder to inherit essential feature representations from the original SAM encoder.
The optimization process also incorporates progressive compression techniques, where training is performed on a subset of the SA-1B dataset~\cite{kirillov2023segment}, using only a small fraction of the original data.
This approach significantly reduces computational requirements while preserving the effectiveness of the model.
Finally, instead of relying solely on transformer layers, the new lightweight encoder integrates convolutional blocks with inverted residuals in its early stages.
This hybrid architecture enables efficient local feature extraction while maintaining the global context modeling capabilities of the Transformers.
\ours{} incorporates all these optimizations and utilizes the MobileSAM image encoder as it is.

Formally, the \ours{} encoder can be represented as a function $f_{\theta{}}$, parameterized by $\theta{}$, that takes an input image $I\in{}\mathcal{R}^{H\times{}W\times{}C}$, and maps it to an embedding $E\in{}R^{d}$.
This process can be formulated as:
\begin{equation}
E=f_{\theta{}}(I)\;,
\end{equation}
where $H$, $W$, and $C$, denote the height, width, and number of channels of the input image, respectively, and $d = 3$ is the dimensionality of the output embedding.

\subsection{The Anomaly Score Prediction (ASP) Sub-Network} \label{sec:asp_subnetwork}
To perform anomaly detection, the image embedding $E$ is processed through a series of $N_l$ fully connected linear layers.
Let $g_{\phi{}}$ represent this transformation, parameterized by $\phi{}$, mapping the embedding $E\in{}\mathbb{R}^{d}$ to an anomaly score $s\in{}[0,1]$.
This can be expressed as:
\begin{equation}
s = g_{\phi}(E)\;,
\end{equation}
where $g_{\phi{}}$ consists of multiple linear layers followed by the Rectified Linear Unit (ReLU) activation function.

The model is trained using a Weighted Binary Cross-Entropy (WBCE) loss function, defined as:  
\begin{equation}
\mathcal{L}_{\text{WBCE}} = - \frac{1}{N} \sum_{i=1}^{N} w_i \left[ y_{i} \log \sigma(s_{i}) + (1 - y_{i}) \log (1 - \sigma(s_{i})) \right]\;, 
\end{equation}
where $N$ is the number of samples, $y_i\in{}{0,1}$ represents the ground truth label (0 for normal, 1 for anomalous), and $s_{i}$ is the predicted anomaly score for the $i$-th sample.
The Sigmoid activation function is denoted by $\sigma$.
The weight $w_i$ is applied to the positive class and is calculated as the ratio of negative to positive samples, to ensure that the model adequately compensates for the class imbalance.
Parameters $\phi{}$ are optimized to minimize this loss, ensuring that normal samples are assigned low scores while anomalous samples receive high scores.

\section{Experiments} \label{cha:experiments}

\subsection{Experimental Details} \label{sec:experimental_details}

\paragraph{\textbf{Datasets.}}
\ours{} has been evaluated on the two well-known industrial anomaly detection datasets, \emph{i.e.}, MVTec-AD~\cite{bergmann2019mvtec}, and ViSA~\cite{zou2022spot}.
The MVTec AD contains 5,354 images distributed in 15 object categories, while the VisA is comprised of 10,821 images distributed in 12 object categories.
The anomalies present in both datasets are of various types, shapes, and scales.

\paragraph{\textbf{Evaluation metrics.}}
In accordance with previous work~\cite{zavrtanik2021draem,gudovskiy2022cflow,yao2023focus}, the anomaly detection performance is measured via the \emph{Image-level Area Under the Receiver-Operator Curve (I-AUROC)}.
Furthermore, according to~\cite{li2024sam}, we also report the results in terms of \emph{inference time} and \emph{number of parameters}.

\paragraph{\textbf{Implementation details.}}
All the code is implemented in PyTorch~\cite{paszke2019pytorch}.
On the MVTec-AD dataset, we train our models for 35 epochs, with a learning rate of $1\times{}10^{-2}$, using Adam~\cite{kingma2014adam} as an optimizer on an NVIDIA RTX 4090, using $N_l=2$ fully connected layers to process image embedding $E$.
Instead, on the ViSA dataset, we train our model for 50 epochs and use $N_l=3$ fully connected layers; the other hyperparameters remain the same as in the MVTec-AD dataset.

\subsection{Anomaly Detection on the MVTec-AD Dataset}
\begin{table}[t!]
    \centering
    \caption{Anomaly detection in term of I-AUROC (\%) on the MVTec-AD dataset~\cite{bergmann2019mvtec}.
    The inference time is measured in\,ms, while the parameter count is expressed in millions.
    In \textcolor{gold_dark}{gold}, the best results. 
    In \textcolor{silver_dark}{Silver} is the second best.
    Competitors' results are from~\cite{li2024sam}.}
    \resizebox{\textwidth}{!}{\begin{tabular}{lccccc}
\toprule
\textbf{Class} & \makecell{\textbf{DRAEM}~\cite{zavrtanik2021draem}\\\scriptsize{ICCV 2021}}
               & \makecell{\textbf{CFLOW}~\cite{gudovskiy2022cflow}\\\scriptsize{WACV 2022}}
               & \makecell{\textbf{PatchCore}~\cite{roth2022towards}\\\scriptsize{CVPR 2022}}
               & \makecell{\textbf{RD4AD}~\cite{deng2022anomaly}\\\scriptsize{CVPR 2022}}
               & \makecell{\ours{}\\\scriptsize{(\textbf{ours})}}\\
\midrule
Carpet     & 96.90 & 97.60 & 99.10 & 98.70 & 100   \\
Grid       & 99.90 & 98.10 & 97.30 & 100   & 99.87 \\
Leather    & 100   & 99.90 & 100   & 100   & 100   \\
Tile       & 100   & 97.10 & 99.30 & 99.70 & 100   \\
Wood       & 99.50 & 98.70 & 99.60 & 99.50 & 96.49 \\
Bottle     & 98.00 & 99.90 & 100   & 100   & 98.82 \\
Cable      & 90.90 & 97.60 & 99.90 & 96.10 & 100   \\
Capsule    & 90.90 & 97.00 & 98.00 & 96.10 & 99.47 \\
Hazelnut   & 100   & 100   & 100   & 100   & 99.17 \\
Metal nut  & 100   & 98.50 & 99.90 & 100   & 92.43 \\
Pill       & 97.10 & 96.20 & 97.50 & 98.70 & 100   \\
Screw      & 98.70 & 93.10 & 98.20 & 97.80 & 99.63 \\
Toothbrush & 100   & 98.80 & 100   & 100   & 100   \\
Transistor & 100   & 92.90 & 99.90 & 95.50 & 99.74 \\
Zipper     & 100   & 97.10 & 99.50 & 97.90 & 100 \\
\midrule
\rowcolor{blue!10} Average    & 97.60 & 97.50 & 99.20 & 98.70 & 99.10 \\
\midrule
\rowcolor{green!10} Latency           & 159 & 37    & 180    & 28     & \bm1{5} \\
\rowcolor{green!10} Parameters        & 97  & 94.70 & 186.55 & 150.64 & \bm1{11.53} \\
\midrule
\midrule
\textbf{Class} & \makecell{\textbf{SimpleNet}~\cite{liu2023simplenet}\\\scriptsize{CVPR 2023}}
               & \makecell{\textbf{RD++}~\cite{tien2023revisiting}\\\scriptsize{CVPR 2023}}
               & \makecell{\textbf{FOD}~\cite{yao2023focus}\\\scriptsize{ICCV 2023}}
               & \makecell{\textbf{STLM}~\cite{li2024sam}\\\scriptsize{ACM MCCA 2025}} 
               & \makecell{\ours{}\\\scriptsize{(\textbf{ours})}}\\
\midrule
Carpet     & 99.70 & 100   & 100   & 99.48 & 100   \\
Grid       & 99.70 & 100   & 100   & 95.57 & 99.87 \\
Leather    & 100   & 100   & 100   & 100   & 100   \\
Tile       & 99.80 & 99.70 & 100   & 100   & 100   \\
Wood       & 100   & 99.30 & 99.10 & 100   & 96.49 \\
Bottle     & 100   & 100   & 100   & 100   & 98.82 \\
Cable      & 99.90 & 99.20 & 99.50 & 98.98 & 100   \\
Capsule    & 97.70 & 99.00 & 100   & 98.69 & 99.47 \\
Hazelnut   & 100   & 100   & 100   & 100   & 100   \\
Metal nut  & 100   & 100   & 100   & 100   & 92.43 \\
Pill       & 99.00 & 98.40 & 98.40 & 98.20 & 100   \\
Screw      & 99.00 & 98.90 & 96.70 & 99.47 & 99.63 \\
Toothbrush & 99.70 & 100   & 94.40 & 99.06 & 100   \\
Transistor & 100   & 98.50 & 100   & 97.58 & 99.74 \\
Zipper     & 99.90 & 98.60 & 99.70 & 99.74 & 100   \\
\midrule
\rowcolor{blue!10} Average    & \bm1{99.57} & \bm2{99.44} & 99.20 & 99.05 & 99.10 \\
\midrule
\rowcolor{green!10} Latency           & 39     & 30     & 103   & \bm2{20}    & \bm1{5} \\
\rowcolor{green!10} Parameters        & 52.88  & 154.87 & 28.83 & \bm2{16.56} & \bm1{11.53} \\
\bottomrule
\end{tabular}
}
    \label{tab:mvtec_results}
\end{table}

The results for image-level anomaly detection on the MVTec-AD dataset are shown in Table~\ref{tab:mvtec_results}.
\ours{} significantly outperforms several recent approaches while maintaining a low computational footprint.

Specifically, compared to STLM~\cite{li2024sam} (our main competitor), which achieves an accuracy of 99.05\%, \ours{} improves the I-AUROC performance while reducing the number of parameters from 16.56\,M to 11.53\,M, and the inference time from 20\,ms to 5\,ms, setting the new state-of-the-art for these two metrics.
Furthermore, \ours{} also achieves competitive I-AUROC compared to SimpleNet~\cite{liu2023simplenet} (the architecture with the highest I-AUROC), while having 78\% fewer parameters and a speedup of more than 7$\times{}$.
RD++~\cite{tien2023revisiting} achieves the second highest I-AUROC (99.44\%) but requires 154.87\,M parameters, while \ours{} achieves a remarkably close accuracy with a 6$\times{}$ reduction in the parameter count.
Similarly, PatchCore~\cite{roth2022towards}, achieves 99.20\% of I-AUROC but requires 186.55\,M parameters.
With respect to this, \ours{} achieves a slightly slower average I-AUROC (99.10\%) but is fully justified by significantly fewer parameters (11.53\,M vs. 186.55\,M) and a lower latency (5\,ms vs. 180\,ms).

\subsection{Anomaly Detection on the ViSA Dataset}
\begin{table}[t!]
    \centering
    \caption{Anomaly detection in term of I-AUROC (\%) on the ViSA dataset~\cite{zou2022spot}.
    In \textcolor{gold_dark}{gold}, the best results. 
    \textcolor{silver_dark}{Silver} the second best.
    Competitors' results are from~\cite{yao2024glad}.}
    \resizebox{\textwidth}{!}{\begin{tabular}{lccccc}
\toprule
\textbf{Class} & \makecell{\textbf{DRAEM}~\cite{zavrtanik2021draem}\\\scriptsize{ICCV 2021}}
               & \makecell{\textbf{PatchCore}~\cite{roth2022towards}\\\scriptsize{CVPR 2022}}
               & \makecell{\textbf{RD4AD}~\cite{deng2022anomaly}\\\scriptsize{CVPR 2022}}
               & \makecell{\textbf{SimpleNet}~\cite{liu2023simplenet}\\\scriptsize{CVPR 2023}}
               & \makecell{\ours{}\\\scriptsize{(\textbf{ours})}}\\
\midrule
Candle     & 89.6 & 98.7 & 96.2 & 96.9 & 98.1 \\
Capsules   & 89.2 & 68.8 & 91.8 & 89.5 & 69.7 \\
Cashew     & 88.3 & 97.7 & 98.7 & 94.8 & 90.7 \\
Chewinggum & 96.4 & 99.1 & 99.3 & 100  & 88.3 \\
Fryum      & 94.7 & 91.6 & 96.9 & 96.6 & 99.1 \\
Macaroni1  & 93.9 & 90.1 & 98.7 & 97.6 & 95.5 \\
Macaroni2  & 88.3 & 63.4 & 91.4 & 83.4 & 89.7 \\
Pcb1       & 84.7 & 96.0 & 96.7 & 99.2 & 96.5 \\
Pcb2       & 96.2 & 95.1 & 97.2 & 99.2 & 87.2 \\
Pcb3       & 97.4 & 93.0 & 96.5 & 98.6 & 95.9 \\
Pcb4       & 98.9 & 99.5 & 99.4 & 98.9 & 99.1 \\
Pipe fryum & 94.7 & 99.0 & 99.6 & 99.2 & 97.8 \\
\midrule
\rowcolor{blue!10} Average    & 92.4 & 91.0 & \bm1{96.9} & \bm2{96.2} & 92.3 \\
\midrule
\rowcolor{green!10} Latency           & 159 & \bm2{37}    & 180    & 39           & \bm1{5} \\
\rowcolor{green!10} Parameters        & 97  & 94.70       & 186.55 & \bm2{52.88}  & \bm1{10.55} \\
\bottomrule
\end{tabular}
}
    \label{tab:visa_results}
\end{table}

The results for image-level anomaly detection on the ViSA dataset are shown in Table~\ref{tab:visa_results}.
Due to space limitations, not all competitors listed in Table~\ref{fig:fig_1} are included; we have selected the top performers while maintaining the time frame mentioned above.
STLM~\cite{li2024sam} is not reported as it only reports overall ViSA results (93.83\%), without providing class-specific performance.

Even in this case, \ours{} achieves the best inference speed and lowest parameter count among all methods compared.
Although \ours{} ranks fourth in overall I-AUROC performance (92.3\%), it outperforms the top two models in several individual classes.
However, its performance is notably affected by the ``Capsules'' class, which remains a key limitation and will be the focus of future research.

\subsection{Deployment of KairosAD on the ICE Lab at the University of Verona}
\begin{figure*}[t!]
    \centering
    \includegraphics[width=\linewidth]{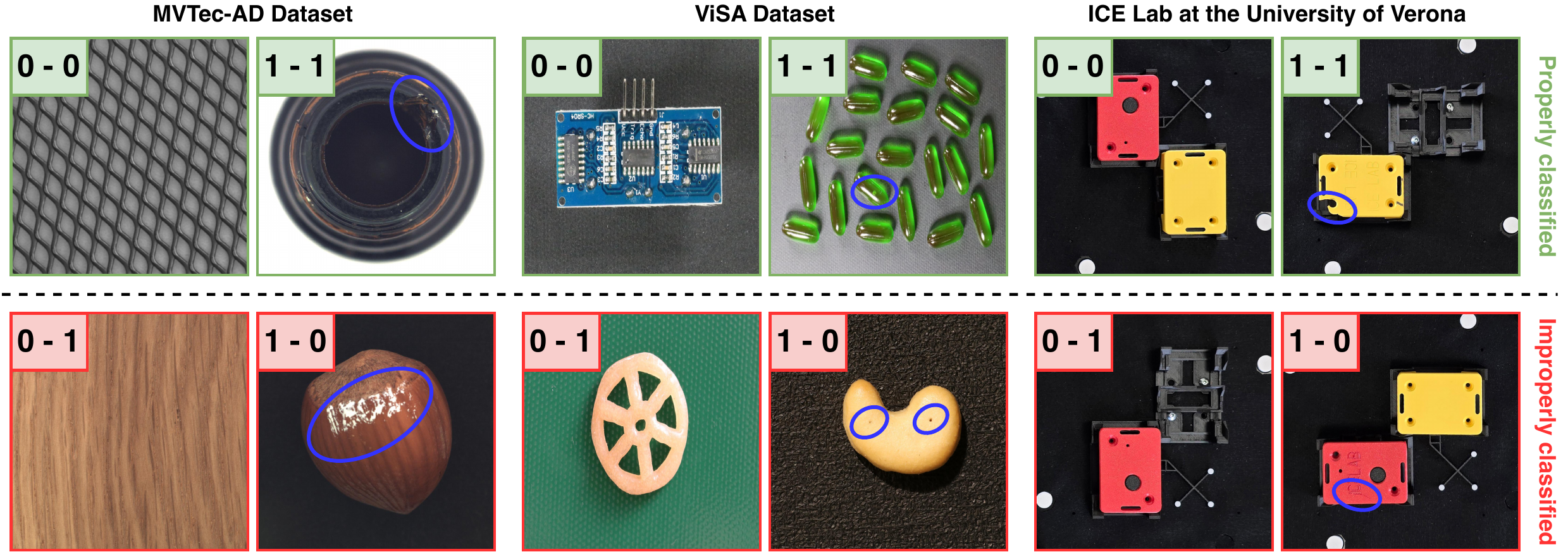}
    \caption{Qualitative results of \ours{} in correctly detecting (and not detecting anomalies) on the MVTec-AD, ViSA, and the real production line of the ICE Lab at the University of Verona.
    For each image, we report the [True Label] - [Prediction] on the top left (where 0 stands for normal and 1 stands for abnormal).
    The anomalies are circled in blue.}
    \label{fig:fig_3}
\end{figure*}

We deployed \ours{} on various embedded devices, including NVIDIA Jetson NX, and NVIDIA Jetson AGX, achieving inference times of 211\,ms and 218\,ms, respectively.
These devices were also the ones we installed and tested on the real production line of ICE Lab at the University of Verona.

In Figure~\ref{fig:fig_3}, we present the qualitative results of \ours{} deployed on the production line, together with the results of other datasets, further validating the robustness and versatility of the model in different industrial settings.

\section{Conclusions} \label{cha:conclusions}

In this article, we present \ours{}, an efficient image-based anomaly detection model optimized for real-time deployment on embedded devices.
Using MobileSAM, \ours{} achieves a balance of high accuracy and low computational cost, outperforming the state-of-the-art methods in inference speed and parameter efficiency.
We demonstrated its effectiveness on multiple industrial datasets and successfully deployed it on different embedded devices on a real production line.
This work provides a practical solution for anomaly detection in resource-constrained environments, with potential for broader applications in industrial settings.

\bibliographystyle{splncs04}
\bibliography{01_main}

\end{document}